\documentclass[journal]{IEEEtran}

\usepackage{graphicx}
\usepackage{epstopdf}
\usepackage{subfigure}
\usepackage{subcaption}
\usepackage[numbers,sort&compress]{natbib}

\usepackage{caption3}
\usepackage{amssymb}
\usepackage{amsmath}

\usepackage{amsfonts}

\usepackage{bbding}
\usepackage{array,diagbox,siunitx}
\usepackage{cases}
\usepackage{float}
\usepackage{multirow}
\usepackage{mathrsfs}
\usepackage{algorithmic}
\usepackage[ruled]{algorithm2e}
\usepackage{rotating}
\usepackage{balance}
\usepackage{threeparttable}
\usepackage{color}
\usepackage{colortbl}
\definecolor{mygray}{gray}{.9}
\usepackage{makecell}
\usepackage{pdflscape}
\usepackage{siunitx}
\usepackage{stfloats}
\usepackage{verbatim}
\usepackage{setspace}
\usepackage{threeparttable}
\usepackage{supertabular,booktabs}
\usepackage{rotating}
\usepackage{supertabular}
\usepackage{pifont}
\usepackage{xcolor,pict2e}% to allow any radius

\usepackage{tikz}
\usepackage{booktabs}
\usepackage{footnote}
\usepackage[pagewise]{lineno}
\usepackage{balance}

\usepackage{hyperref}
\usepackage{ulem}

\newcommand{\etal}{\textit{et al.}}
\newcommand{\ie}{\textit{i.e.}}
\newcommand{\eg}{\textit{e.g.}}

\begin{document}
\title{Towards A Robust Group-level Emotion Recognition via Uncertainty-Aware Learning}
\author{Qing Zhu,
        Qirong Mao*,~\IEEEmembership{Member,~IEEE},
        Jialin Zhang,
        Xiaohua Huang*,~\IEEEmembership{Senior Member,~IEEE},
        Wenming Zheng,~\IEEEmembership{Senior Member,~IEEE}
        % <-this % stops a space
        % <-this % stops a space
\thanks{{Q. Zhu, Q. Mao, J. Zhang are with the School of Computer Science and Communication Engineering, Jiangsu University, Zhenjiang, Jiangsu, China. (E-mail: zhuqing@stmail.ujs.edu.cn, mao\_qr@ujs.edu.cn,  2212108042@stmail.ujs.edu.cn)}}

\thanks{X. Huang with Oulu School and School of International Education, Nanjing Institute of Technology, Jiangsu, China and also with the Key Laboratory of Child Development and Learning Science (Southeast University), Ministry of Education, Southeast University, Nanjing 210096, China.
(E-mail: xiaohuahwang@gmail.com)}

\thanks{W. Zheng is with the Key Laboratory of Child Development and Learning Science (Southeast University), Ministry of Education, Southeast University, Nanjing 210096, China and also with the School of Biological Science and Medical Engineering, Southeast University, Nanjing 210096, Jiangsu, China.
(E-mail: wenming\_zheng@seu.edu.cn)}

\thanks{This research was in part supported by the Basic Science (Natural Science) Research Project of Higher Education Institutions in Jiangsu Province under Grant 24KJA520003 and 333 high-level talents in Jiangsu Province (2024), in part by the National Natural Science Foundation of China under Grant No. 62076122, 62176106, U2003207, in part by the Fundamental Research Funds for the Central Universities under Grant 2242024k30027.}
\thanks{\textsuperscript{*} Co-Corresponding author: Xiaohua Huang and Qirong Mao.}

}

\markboth{IEEE Transactions on Affective Computing, VOL. x, NO. x, 2025}%
{Shell \MakeLowercase{\textit{et al.}}: Bare Demo of IEEEtran.cls for IEEE Journals}

%
% Single address.
% ---------------
%\address{}
\maketitle
\begin{abstract}
Group-level emotion recognition (GER) is an inseparable part of human behavior analysis, aiming to recognize an overall emotion in a multi-person scene. However, the existing methods are devoted to combing diverse emotion cues while ignoring the inherent uncertainties under unconstrained environments, such as congestion and occlusion occurring within a group. Additionally, since only group-level labels are available, inconsistent emotion predictions among individuals in one group can confuse the network. In this paper, we propose an uncertainty-aware learning (UAL) method to extract more robust representations for GER. By explicitly modeling the uncertainty, we adopt stochastic embedding sourced from a Gaussian distribution instead of deterministic point embedding. It helps capture the probabilities of emotions and facilitates diverse inferences. Additionally, we adaptively assign uncertainty-sensitive scores as the fusion weights for individuals' faces within a group. Moreover, we developed an image enhancement module to evaluate and filter samples, strengthening the model's data-level robustness against uncertainties. The overall three-branch model, encompassing face, object, and scene components, is guided by a proportional-weighted fusion strategy and integrates the proposed uncertainty-aware method to produce the final group-level output. Experimental results demonstrate the effectiveness and generalization ability of our method across three widely used databases.

\end{abstract}

\IEEEpeerreviewmaketitle
\section{Introduction}
\label{sec:intro}

\IEEEPARstart{A}UTOMATIC recognition of human emotions has been extensively studied in the field of multimedia computing, encompassing image, audio, text, and video analysis. This research significantly contributes to the understanding of human behavior. Over the past decades, researchers have made substantial progress in individual-level emotion recognition \cite{rouast2019deep,LiD22a}. According to investigations in social sciences~\cite{barsade2015group,barsade2012group}, human beings may alter their reactions and behavior based on their perception of the emotions of those around them. Consequently, group-level emotion recognition (\textbf{GER}) has garnered significant attention in recent years. Unlike individual-level emotion recognition, GER focuses on collectively detecting emotions expressed by groups of people. Moreover, GER has broad societal implications spanning various fields, such as social behavior analysis, public security, and human-robot interactions~\cite{SanchezHTH20,VeltmeijerGH23}. Given the multitude of uncertain emotion cues, learning meaningful and robust representations for GER across the entire scene poses a considerable challenge.

\begin{figure}[t]
\includegraphics[width=\linewidth]{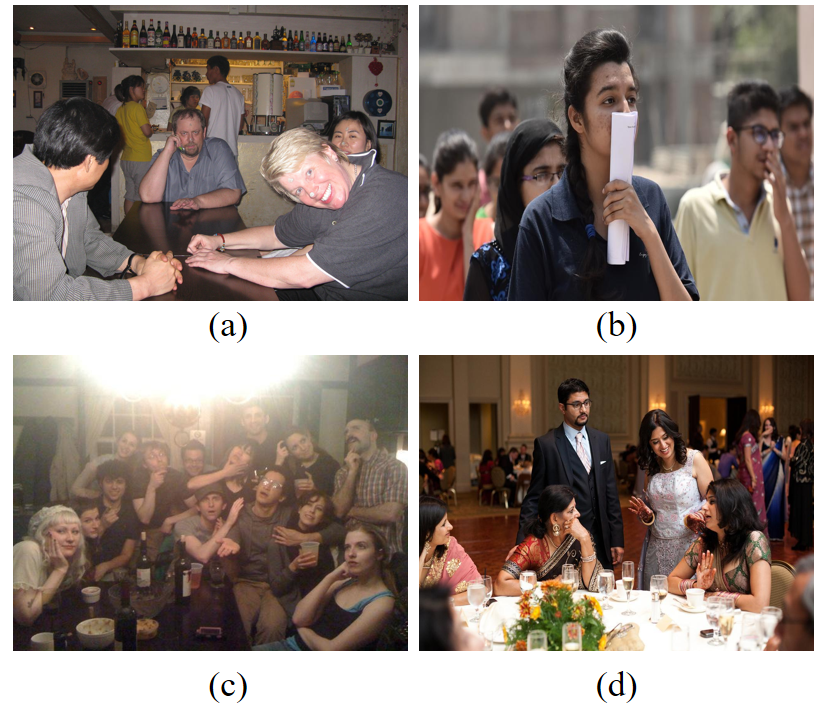}
\caption{Observation and Motivation: Low-quality examples in the GER database contain varying degrees of uncertain information. In (a), (b), and (c), a face is partially obscured due to being blocked by another individual in the same group, while in (b) and (d), faces experience self-occlusion. Robust emotion representations are necessary to assign lower weights to these face samples. Emotion predictions for individuals are ambiguous in both (a) and (d), and emotions of objects with the same semantic information vary in (a) and (c). These factors significantly impact the performance of GER.}
\label{fig:motivation}
\end{figure}

% The group-level emotion label
GER is built upon effectively combining individual-level information and a comprehensive understanding of different compositions. Researchers have explored techniques to adequately model diverse emotion-related features in a group-level image\footnote{For the sake of simplicity, we refer to a “group-level image" as an “image" throughout this text, although it contains more than two persons.} and to efficiently aggregate these features for the purpose of group-level emotion inference \cite{guo2020graph,GuoZPBB18,GuoPB17}. Previous work primarily focused on capturing emotion features from both faces and scenes within the image \cite{huang2019analyzing,fujii2019hierarchical,quach2022non}, as these features can explicitly convey group-level emotion. Some existing studies \cite{wang2022congnn,fujii2020hierarchical} proposed integrating objects at the individual level to fully capture emotional information, resulting in significant improvements in GER. Similarly, our framework integrates scene and individual features (\ie, faces and objects) from global and local perspectives for GER. To elaborate, given an image, we first extract individual and scene emotion features using their respective backbone networks. Subsequently, individual features are employed to generate group-level representations using methods such as arithmetic averaging, voting, Recurrent Neural Networks (RNNs), or attention mechanisms. Finally, a final group-level emotion prediction is produced by fusing all prediction results or refined features from individuals and the scene. 

However, based on our current understanding, nearly all existing GER methods represent group-level emotions deterministically. Specifically, when presented with an image, these methods utilize deep networks to generate deterministic point embeddings. This approach, however, neglects the inherent data ambiguity that arises in realistic scenarios. Such oversight significantly limits the creation of robust emotion representations. Two types of uncertainties are indeed present within the images. The first type encompasses factors like congestion, occlusion, illumination variations, and more. These factors stem from the intrinsic complexity of groups. For instance, as shown in Fig.~\ref{fig:motivation}, mutual occlusion and self-occlusion frequently lead to the absence of individuals' information. Since congestion is pervasive, it closely ties to the fundamental attributes of a group. Regarding the second type, it's important to note that different individuals within the same group might not display identical emotions, despite sharing the same group-level emotion label. As depicted in Fig.~\ref{fig:motivation}(a), the gentleman on the right smiles, while the middle one does not. Similarly, in Fig.~\ref{fig:motivation}(d), the lady exhibits a smiling face, while the bespectacled gentleman's expression remains neutral.  Furthermore, objects with identical semantics might evoke diverse emotions in different groups. Take the example of ``beer",  which appears in both Fig.~\ref{fig:motivation}(a) and Fig.~\ref{fig:motivation}(c), yet bears different emotion labels. These phenomena can potentially contradict the inference of group-level emotions. Treating individuals deterministically would severely compromise the performance of GER due to the influence of these uncertainties. Therefore, mitigating the aforementioned uncertainties surrounding group-level emotions becomes critical.

In this paper, we propose an Uncertainty-Aware Learning (\textbf{UAL}) method to enhance the robustness of emotion representations for GER. Specifically, we introduce a probability distribution to generate a stochastic representation for each individual, departing from the deterministic point embeddings used in existing methods. For the sake of modeling simplicity, we map each individual to a Gaussian distribution in latent space, characterized by mean and variance parameters. The former signifies the feature, while the latter quantifies the uncertainty. Critically, the feature instance of each individual is treated as a random variable originating from a Gaussian distribution. Leveraging this uncertainty modeling, diverse predictions arise due to the inherent stochasticity. This approach mitigates the adverse effects of uncertainty, resulting in more resilient emotion features for GER. Moreover, we estimate the variance by producing allocations of uncertainty-sensitive scores. This leads to adaptive down-weighting of individuals with high scores (indicative of large uncertainty) during the fusion stage of the final GER process. Furthermore, we introduce an image enhancement module to counteract the impact of nearly unrecognizable individuals at the data level. Lastly, to integrate multiple emotion cues and the UAL module into a unified framework, we design a three-branch model comprising face, object, and scene branches. This model can independently infer emotions or combine information using various strategies, harnessing complementary information for group-level emotion inference.

The main contributions of this work are summarized as follows:
\begin{itemize}
    \item We introduce a novel reasoning paradigm that aligns with the Uncertainty-Aware Learning module. This paradigm enables the encoding of latent uncertainty among all individuals and facilitates the learning of more robust representations for GER. Notably, our approach is pioneering in modeling uncertainty across all individuals within a group for the GER task. 
    
    \item To combat uncertainty, we present probability distribution instead of deterministic feature vectors to represent individuals. It facilitates generating diverse emotion representations and formulation of uncertainty-sensitive score allocations, adaptively downweight more uncertain samples during face individual aggregation. Additionally, we filter and enhance face individual samples to further mitigate the adverse effects of uncertainty on GER.
    
    \item We conduct comprehensive experiments on three different group-level emotion databases. Through comparisons with various variants and state-of-the-art methods, our approach's efficacy for GER is convincingly demonstrated.
   
\end{itemize}

The rest of this paper is structured as follows. Section~\ref{sec:related} presents an overview of the related work. Section~\ref{sec:method} details the architecture of the proposed method, including uncertainty modeling module, the image enhancement module, and the proportional-weighted fusion module. Section~\ref{sec:experiment}  describes the experiment setup and the performance evaluation. The conclusion is given in Section~\ref{sec:conclusion}.

\section{Related Work}
\label{sec:related}

\subsection{Group-level Emotion Recognition}
Compared to individual-level emotion recognition, GER involves comprehending complex emotions expressed by multiple individuals. Various efforts have been made to enhance GER performance by harnessing diverse emotion-related information from multiple sources and subsequently aggregating individual features into group-level insights ~\cite{abs-1710-01216,huang2019analyzing,TarasovS18,NagarajanO19}. Among these sources, facial feature learning holds prominence in influencing the inference of the ultimate group-level emotion. This is primarily due to facial expressions serving as the most explicit signals that convey emotional states during human interaction~\cite{mehrabian1981silent}. Facial features have been employed to estimate happiness intensity within a group~\cite{dhall2015automatic,huang2015riesz}. Khan~\etal~proposed a four-stream hybrid network, incorporating a multi-scale face stream to handle variations in face size and exploring distinct global streams to capture scene information~\cite{KhanLCMOT18}. Notably, facial information plays a pivotal role in recognizing strong group emotions such as positive and negative. In the course of GER's evolution and in-depth research, recent studies suggest that, apart from facial features, additional information stemming from group-related factors, including objects and scenes, holds potential for enhancing GER. Fujii~\etal~ adopted a hierarchical classification approach, where facial expression features initially underwent binary classification, followed by incorporation of object and scene information into GER~\cite{fujii2020hierarchical}. Guo~\etal~ devised a Graph Neural Network (GNN) to leverage emotional cues from faces, objects, scenes, and skeletons~\cite{guo2020graph}. The advancements in GER research underscore the increasing recognition of the significance of information beyond facial features, indicating that elements like scene context and object interactions have valuable contributions to make in improving GER accuracy and comprehensiveness.

To integrate diverse individual contributions, certain traditional approaches have employed arithmetic-based methods such as averaging~\cite{surace2017emotion,BalajiO17,abbas2017group} or voting~\cite{fujii2019hierarchical,Garg2019,GuptaACDP18}. Rassadin~\etal~developed a strategy involving multiple classifiers to derive the GER outcome by averaging the facial expressions of individuals, their facial landmarks, and the corresponding scene features~\cite{RassadinGS17}. Dejian~\etal~combined facial features from individuals and inputted them into the fully-connected (FC) layer to create a representation of facial features utilized for group-level emotion recognition~\cite{Dejian2020Group}. With the advent of deep learning, some approaches turned to Recurrent Neural Networks (RNNs) and their variants, such as Long Short-Term Memory (LSTM) and Gated Recurrent Unit (GRU) networks~\cite{bawa2019emotional,WeiZXL0YS17,quach2022non}. Bawa~\etal~utilized an LSTM-based approach to aggregate facial features extracted from various regions of the image~\cite{bawa2019emotional}. Additionally, several methods introduced Graph Neural Networks (GNNs) to leverage discriminative emotion representation and capture correlations among individuals~\cite{guo2020graph,wang2022congnn}. Notably, the recent work by Fujii~\etal~\cite{fujii2020hierarchical} introduced an attention mechanism to assess the relative importance of individuals within a group. A new perspective emerges as Zhang~\etal~present a contrastive learning-based semi-supervised framework for group emotion recognition, aimed at efficiently extracting features from both labeled and unlabeled images~\cite{zhang2022semi}.

Indeed, GER research extends beyond acquiring diverse emotion-related information to manage factors that may not contribute effectively to emotion representation. Thus, our work is equally dedicated to aggregating rich information from individuals (face and object) and scenes. However, the crucial distinction lies in our method's objective, which is to enhance the diversity and robustness of representations by explicitly modeling uncertainty among individuals for GER.

\subsection{Learning with Uncertainties}

The concept of uncertainty learning has garnered significant attention in the computer vision domain due to its effectiveness in learning robust and interpretable features. Within the Bayesian framework, uncertainty learning can be broadly categorized into two types: model uncertainty and data uncertainty. Data uncertainty pertains to the inherent noise present in training data, capturing uncertainties originating from data noise. On the other hand, model uncertainty arises from the lack of knowledge concerning potential noise in model parameters. Numerous tasks have embraced the integration of uncertainty to enhance model robustness and interpretability. This trend is evident across various domains, including face recognition~\cite{WuZHSM22}, semantic segmentation~\cite{FanGJJ22,YangFXRDNAR23}, and ReID tasks~\cite{ZhengLZZZ21,DouWCLW22}.

In visual classification tasks, which align closely with our current objective, prior works predominantly address uncertainty through two primary avenues. One is focused on designing a mechanism to quantify uncertainty to address challenges arising from low-quality samples and label noise. Wang~\etal~proposed the SCN to suppress the impact of uncertainties via a self-attention mechanism to weight each sample with a ranking regularization. She~\etal~employed a method to model latent label distribution of input samples and identify uncertain samples through a cosine similarity learning branch~\cite{she2021dive}. Otherwise, in human-centered visual tasks~\cite{DouWCLW22,ChangLCW20,ZhaoCDQYWM22,ChenLD0L22}, several studies have employed uncertainty-aware learning to model uncertainty, particularly utilizing the method of modeling samples as Gaussian distribution to address the challenges like congestion, occlusion, illumination changes, and inconsistent predictions. Specifically, to handle noisy face images, Chang~\etal~employed a Gaussian distribution estimate for individual face images in the latent space and acquired identity features (mean) while also capturing the uncertainty (variance) associated with the estimated mean~\cite{ChangLCW20}. Dou~\etal~proposed an uncertainty-aware learning method for jointly learning data uncertainty and model uncertainty in a unified network, simultaneously addressing both issues of low-quality samples and inconsistent predictions in the person image retrieval task~\cite{DouWCLW22}. Zhao~\etal~proposed a context-aware contrastive loss to learn more robust representations and introduced data uncertainty learning to distinguish congested and overlapped pedestrians \cite{ZhaoCDQYWM22}. 

Similar with the typical setting of learning in the uncertainty learning area, we employ the  Gaussian sampling for generating stochastic representations and employing KL divergence for regularization.  on this basis, we introduce a reconstruction loss to regularize uncertainty-sensitive scores generated through covariance approximation. This introduction of the reconstruction loss results in a similarity to VAE. To further clarify the differences between our approach and VAEs, it is important to highlight two key distinctions: 1) For Representation Learning: VAEs generate a latent space representation of inputs by encoding them into a distribution, typically Gaussian, and then sampling from this distribution to reconstruct the input. Our method also uses Gaussian sampling, but unlike VAEs, our focus is not on reconstructing the original input. Instead, our goal is to produce a robust emotional representation, where the emphasis is on capturing the variability and uncertainty in group emotion recognition. 2) For Regularization Techniques: In VAEs, the KL divergence is used as a regularization term to ensure that the latent distribution closely matches a prior distribution, aiding in the reconstruction of inputs. Our approach employs KL divergence for regularization, but the crucial difference lies in our unique regularization term, a reconstruction loss term, designed to stabilize training. This term aims to control the instability in training due to uncertainty fluctuations, rather than ensuring the fidelity of reconstructed data to the input.

Actually, nearly all existing GER methods represent group-level emotions deterministically, which leads to a lack of research on uncertainty approximation within the realm of GER. Our approach is inspired by the interaction of the aforementioned aspects. Building upon the generation of stochastic representations using Gaussian distributions, distinct from prior art, we design a mechanism to produce uncertainty scores, enabling adaptive fusion weights in real-world scenarios characterized by high uncertainty, thereby enhancing the robustness of group-level emotion recognition.

% For GER in real-world scenarios, learning with uncertainties becomes pivotal to counteract the effects of low-quality facial images and noisy labels attributed to complex acquisition conditions like illumination, occlusions, and low resolution. In this context, our work stands out as the first to emphasize intrinsic group uncertainties, modeling each individual as a Gaussian distribution. This approach facilitates diverse predictions through probabilistic representations during the inference stage. By incorporating such probabilistic uncertainty modeling into GER, we aim to enhance the diversity and robustness of the emotion recognition process.

\section{Proposed method}
\label{sec:method}

\begin{figure*}[t!]
\centering
\includegraphics[width=\textwidth]{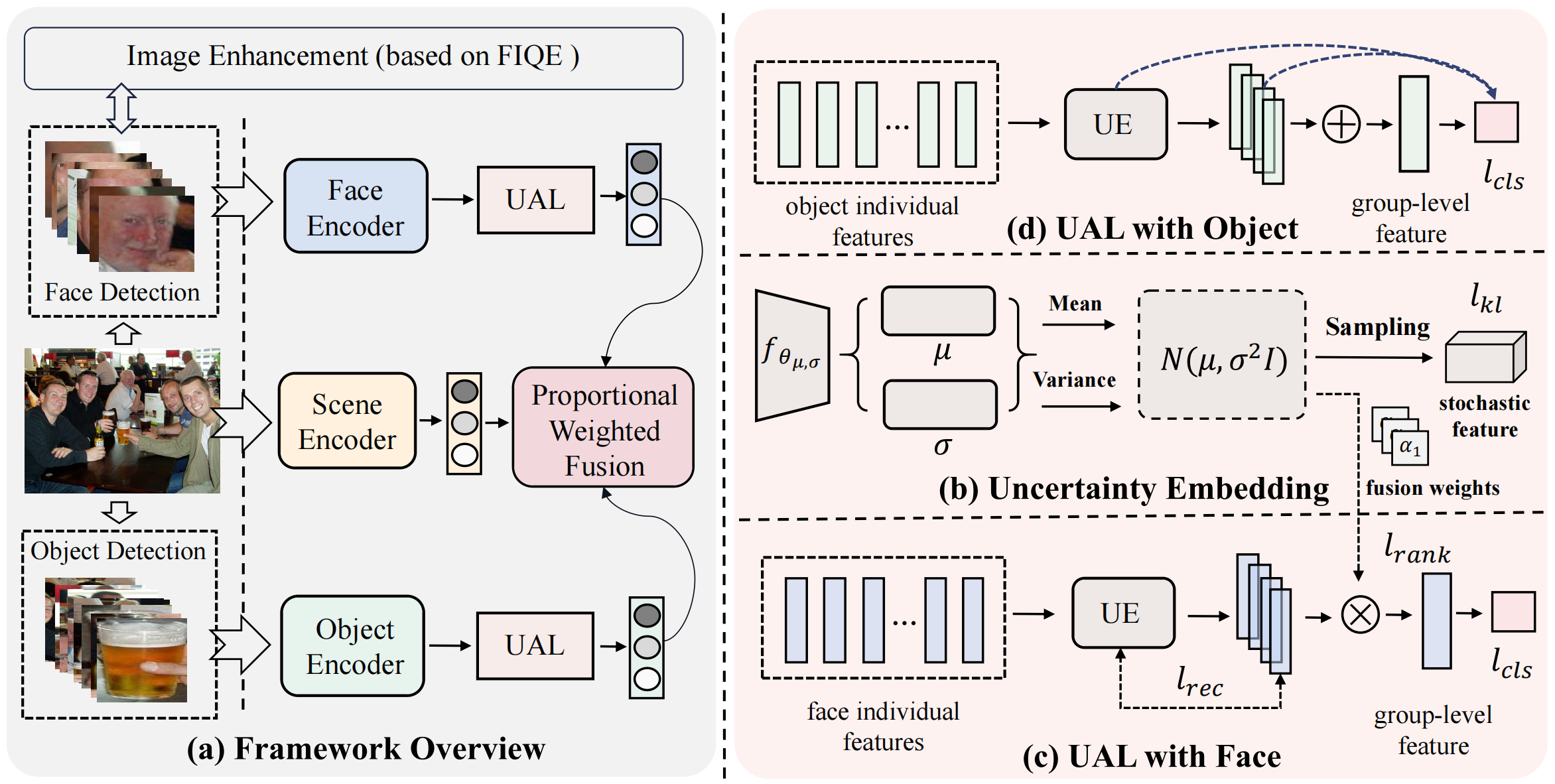} 
\caption{The overview of our proposed method is depicted. The framework of the proposed method is illustrated in (a), incorporating face, object, and scene branches for GER and integrating the UAL module into the face and object branches. Notably, the face branch includes an image enhancement module. The proportional-weighted fusion combines the outputs of the three branches to provide the final group-level prediction. The UAL module is shown in (b-d). Uncertainty embedding (UE) in (b) represents each individual using stochastic embedding rather than the conventional point embedding. (c) and (d) correspond to the modeling of uncertainty with UE incorporated into the face and object branches, respectively.}
\label{fig:overall}
\end{figure*}

The goal of the proposed method is to recognize group-level emotion in a crowd scene by aggregating more robust emotion features. With the estimated uncertainty, the representation of each individual is based on probabilistic distribution to infer the group-level emotion.
In this section, we first present all descriptors of our method. Then we emphatically detail an uncertainty-aware learning module and apply it to model faces and objects. Finally, we describe a simple but effective image enhancement module and the proportional-weighted fusion module used to aid in GER.

The proposed framework including the UAL module is illustrated in Fig.~\ref{fig:overall}. The overview is shown in Fig.~\ref{fig:overall}(a), which consists of face-level, object-level, and scene-level branches for GER. Given an image, firstly, the detectors are utilized to generate a set of face and object proposals. A simple image enhancement module is attached to the face branch. Second, the CNN-based feature extractors are used to extract local representations for each face and object on face and object branches and global representation in the scene branch, respectively. Next, the extracted individual features are mapped into the corresponding distributions by using the UAL module. Then, in the inference stage, the Monte Carlo sampling operation is used to obtain the diversity prediction of the individual. After that, the individual features output from the UAL module are aggregated as the corresponding final group-level predictions. Finally, the predicted emotion categories of the three branches are fused by a proportional-weighted fusion module to refine the final prediction of the input image. 

\subsection{Feature Extractor}
To recognize the group-level emotion, we solve it by recognizing individual emotion and scene emotion simultaneously. For individual emotion recognition, we need to drive the detector to obtain face and object images as a premise. In this work, we adopt MTCNN \cite{zhang2016joint} as the detector to obtain faces, which is a multi-task cascaded convolutional network widely used in face and landmark detection. For the acquisition of objects, the Faster R-CNN \cite{RenHG017} is utilized to generate a set of object proposals and train on the MSCOCO dataset. 

To capture not only the local individual representation but the global scene representation, we utilize three distinct encoders to severally extract the features for each branch. For the face branch, we select ResNet18 pre-trained on the MS-Celeb-1M \cite{guo2016ms} dataset as the encoder as it was recently efficiently used for the facial expression recognition tasks \cite{WangPYL020,zhang2021relative,yan2022mitigating} while achieving remarkable results on uncertainty estimation benchmarks. For object and scene branches, we adopt the VGG19 network pre-trained with the ImageNet dataset as the encoder. We define the output of three feature extractor corresponding branches as $x_{i}^{f}$, $x_{i}^{o}$ and $x^{s}$, which represent the extracted face feature of the $i$th cropped face image, object feature of the $i$th cropped object image, and scene feature of the corresponding whole image, respectively.

\subsection{Uncertainty Modeling}
The UAL module is illustrated in Fig.~\ref{fig:overall}(b-d). Conventionally, the individual feature obtained by the feature extractor is represented as a deterministic point in space. However, it is difficult to estimate an accurate point embedding for the individuals affected by uncertain factors, which is attributed to the complexity of the GER datasets collection scenarios. Furthermore, only a single group-level label is available in the GER task, which means that every individual represents the common emotion category in a group. Nevertheless, individuals in a group may perform emotion with different forms and may spring up several possible emotion categories, which are reflected by the uncertainty, in other words. The conventional GER methods cannot naturally express the uncertainty and distribution of individuals and are unable to effectively quantify the diversity of emotion prediction. Therefore, UAL is proposed to address this issue. UAL is implanted in the face and object branches to reduce the interference of uncertainty for more robust feature learning. 

\noindent\textbf{Uncertainty Embedding.} To explicitly represent the individual feature and the uncertainty simultaneously, the individual feature is modeled as a multivariate Gaussian distribution. In particular, we define the representation $z_{n}$ in latent space of $n$-th individual $x_{n}^{I}$ (\ie, $x_{i}^{f}$, $x_{i}^{o}$) as a Gaussian distribution,
\begin{eqnarray}
P(z_{n}|x_{n}^{I}) = N(z_{n}; \mu_{n}, \sigma_{n}^{2}I),
\end{eqnarray}
where $\mu_{n}\in R^{D}$ and $\sigma_{n}^{2}\in R^{D}$ represent mean vector and diagonal covariance matrix, respectively. $D$ is the individual representation length. The $\mu_{n}$ and $\sigma_{n}^{2}$ are input-dependent predicted by two separate feature projectors by:
\begin{eqnarray}
\mu_{n} = f_{\theta_{\mu}}(z_{n}),
\label{eqn:fmu}
\end{eqnarray}
\begin{eqnarray}
\sigma_{n} = f_{\theta_{\sigma}}(z_{n}),
\label{eqn:fsigma}
\end{eqnarray}
where $\theta_{\mu}$ and $\theta_{\sigma}$ denote corresponding probabilistic parameters respectively \textit{w.r.t.} output $\mu_{n}$ and $\sigma_{n}$. The predicted Gaussian distribution is a diagonal multivariate normal distribution. The $\mu_{n}$ can be regarded as an individual feature, while $\sigma_{n}$ refers to the intensity of the embedding variance that represents the uncertainty of the individuals. Now, the representation of each individual serves as a stochastic instead of a deterministic point embedding sampled from $N(z_{n}; \mu_{n}, \sigma_{n}^{2}I)$. However, the sampling operation is not differentiable for training these probabilistic parameters through the gradient descent, we adopt the re-parameterization trick \cite{KingmaW13} to ensure the operation of back-propagation. To further capture the diversity and uncertainty present in the underlying data distribution, we utilize Monte-Carlo sampling to obtain the final generated stochastic feature representations $z^{*}_{n}$:
\begin{eqnarray}
z^{*}_{n} = \frac{1}{M} \sum_{m=1}^M (\mu_{n} + \epsilon_{m}\sigma_{n}), \epsilon_{m} \sim N(0, I),
\label{eqn:parameterization}
\end{eqnarray}
where $\epsilon_{m}$ is a random number sampled from a normal distribution, $M$ is the number of samples drawn from the Gaussian distribution $N(0, I)$. The representations $z^{*}_{n}$ obtained through Monte Carlo sampling are generated by sampling from a Gaussian distribution, with each sampling iteration producing distinct feature representations. Indeed, during the training phase, the $\mu_{n}$ and $\sigma_{n}$ learned by the $f_{\theta_{\mu}}(*)$ and $f_{\theta_{\sigma}}(*)$ in Eq.~\ref{eqn:fmu} and Eq.~\ref{eqn:fsigma} while fixed during the testing phase, as the model parameters have reached their optimal values through training. The repetitive sampling process, yielding diverse representations, aids in quantifying uncertainty, and addressing inherent uncertainties in the data, and enhances the model's robustness. 
% \begin{eqnarray}
% z^{*}_{n} = \mu_{n} + \epsilon_{n}\sigma_{n}, \epsilon_{n} \sim N(0, I),
% \label{eqn:parameterization}
% \end{eqnarray}

\noindent\textbf{Uncertainty-sensitive score.} Based on Eq.~\ref{eqn:parameterization}, we obtain the random variable $z_{n}^{*}$ as a stochastic representation sample instead of the $\mu_{n}$ sampling from the original distribution. The uncertainty-aware learning module is proposed primarily to decrease the disturbance of the uncertain individuals in the image. Hence, in the face branch, we formulate the uncertainty-sensitive score as the source of the weight for the corresponding face individuals. Specifically, the uncertainty-sensitive score is computed by the harmonic mean of the Hadamard product of the estimated variance $\sigma_{n}$ and random noise $\epsilon$. We denote the uncertainty-sensitive score as $s_{n}$, which acquired by:
\begin{eqnarray}
s_{n} = \frac{D}{\sum_{d=1}^{D} \frac{1}{\sigma_{n,d}\epsilon_{n,d}}},
\end{eqnarray}
where $\sigma_{n,d}$ and $\epsilon_{n,d}$ represent the $d$-th compositions of $\sigma_{n}$ and $\epsilon_{n}$, respectively. Hence, a face individual with higher $s_{n}$ generally corresponds to the larger uncertainty and the other way around.

Once the uncertainty-sensitive score of each face individual is estimated, it is regarded as the criterion of weight. To be specific, we project the uncertainty-sensitive score $s_{n}$ to $\alpha_{n} = \beta_{n} s_{min} + (1-\beta_{n}) s_{max}$, where $\beta_{n} = \frac{s_{n}-s_{min}}{s_{max}-s_{min}}$ and the $s_{min}$ and $s_{max}$ represent the maximum and minimum value of $s_{n}$, respectively. Here, we note $\alpha_{n}$ as the importance scalar for each face individual in a group. Hence, our model can adaptively assign weights to every face individual, down-weighting the individual with the high uncertainty-sensitive score. 

The weighted group-level features in the face branch can be expressed as:
\begin{eqnarray}
x_{group}^{f} = \frac{\sum_{n}z^{*}_{n} \alpha_{n}}{\sum_{n} \alpha_{n}},
\end{eqnarray}
where $x_{group}^{f}$ are the final group-level representations in the
face branch. Apparently, the importance scalar $\alpha_{n}$ plays a role similar to the attention mechanism, enabling the group-level representation is not disturbed by individuals with large uncertainty to a great extent. 

Since the object individuals in a group cannot express emotions as intuitively as human faces and to better compare object individuals with the same semantic information between different groups, we directly predict the emotion of a single sampled object individual and average the sum of all object individual predictions as the group-level emotion prediction.

\noindent\textbf{Uncertainty-aware Loss.} Since $x_{group}^{f}$ is the final group-level representation in the face branch, we feed it to a classifier to minimize the following softmax loss, which is formulated as,
\begin{eqnarray}
L_{cls}^{fa} = -\frac{1}{N} \sum_{n=1}^N \log \frac{e^{W_{y_{i}}x_{group}^{f}}}{\sum_{c}^{C} e^{W_{c}x_{group}^{f}}},
\end{eqnarray}
where $W_{c}$ is the $c$-th classifier and $C$ is the number of emotion categories. 

For the object branch, we treat the $\mu_{n}$ as the original deterministic representation and feed it into the classifier along with the sampled stochastic representation $z_{n}^{*}$ to greatly enrich the semantic information in the object branch. The classification loss is formulated as:
\begin{align}
\nonumber L_{cls}^{ob} & = \lambda_{1} (-\frac{1}{N} \sum_{n=1}^N \log \frac{e^{W_{y_{i}}\mu_{n}}}{\sum_{c}^{C} e^{W_{c}\mu_{n}}}) \\
 & + (1-\lambda_{1})(-\frac{1}{N} \sum_{n=1}^N \log \frac{e^{W_{y_{i}}z_{n}^{*}}}{\sum_{c}^{C} e^{W_{c}z_{n}^{*}}}).
\end{align}

Nevertheless, only the $L_{cls}$ series is employed to constrain the model for classification that easily falls into the trivial solution and reverts our distribution-based embedding back into the deterministic embedding. Hence, it is necessary to constrain $\epsilon$ to avoid the trivial solution by outputting negligible uncertainties. This problem can be alleviated by introducing the regularization term KL divergence during the optimization, it explicitly bounds the learned distribution $N(\mu_{n}, \sigma_{n}^{2}I)$ from the normal distribution $N(0, I)$. This KL divergence term is:
\begin{align}
\nonumber L_{kl} & = KL[N(z|\mu , \sigma)||N(\epsilon|0,I)] \\
 & = -\frac{1}{2N} \sum_{n=1}^N \sum_{d=1}^D (1+log\sigma_{n,d}^{2}-\mu_{n,d}^{2}-\sigma_{n,d}^{2}).
\end{align}

Furthermore, to explicitly constrain the importance scalar of each face individual, we sort the $\alpha_{n}$ in a high-to-low order. Then, similar to \cite{WangPYL020} the sorted face individuals are divided into two groups of high and low importance according to a ratio $\beta$. A margin is used to ensure that the average values of the two groups are maintained in their present size order. Here, the rank regularization loss can be formulated as:
\begin{eqnarray}
    L_{rank} = max(0, \delta_{1} - (\alpha_{H} - \alpha_{L})).
\end{eqnarray}

Inspired by \cite{ChenLD0L22}, we introduce a regularization loss, denoted as $L_{rec}$, to mitigate excessive oscillations in uncertainty-sensitive scores due to the approximation of variance, which may lead to unstable model training. Specifically, $L_{rec}$ is computed as the difference between the original face individual feature $\mu_{n}$ and the sampling $z_{n}^{*}$, formally defined as $L_{rec} = ||z_{n}^{*} - \mu_{n}||_{1}$. The goal of $L_{rec}$ is to preserve the content information of the original feature $\mu_{n}$ to the greatest extent possible.

It is important to clarify that while our method incorporates a regularization approach and a loss component that may resemble those in VAE, there are fundamental differences in their applications. Our model's primary objective is not to reconstruct the input from its latent representation. Instead, there are two main purposes for designing the $L_{rec}$: One is to serve as a regularized term to mitigate instabilities caused by fluctuations in uncertainty. Another is to learn a more stable and deterministic representation of the input data. We aim to leverage the benefits of uncertainty-aware learning \cite{DouWCLW22, ChangLCW20} while ensuring stability and convergence in the training process.

To sum it up, the final loss functions for training the network with the joint loss function of uncertainty and recognition in the face and object branches are severally formulated as follows:
\begin{eqnarray}
    L_{fa} = L_{cls}^{fa} + \lambda_{2} L_{kl} + \lambda_{3} L_{rank} + \lambda_{4} L_{rec},
    \label{eqn:final_face_loss}
\end{eqnarray}
\begin{eqnarray}
    L_{ob} = L_{cls}^{ob} + \lambda_{2} L_{kl}.
    \label{eqn:final_obj_loss}
\end{eqnarray}
In addition, for the scene branch, the cross-entropy loss is used for the training stage.

\subsection{Image Enhancement Module}

In complex scenarios with diverse people, uncertainties make it hard to get reliable facial data and undermine face detection systems, often leading to low-quality samples entering the face individual branch. To solve this, we developed an image enhancement module. It uses face quality assessment as a preliminary safeguard to suppress and filter out bad face samples at the data level, acting like data augmentation and improving dataset quality. But just filtering isn't enough. Despite removing many poor samples by scores, it can't handle all uncertainties like lighting, individual differences and emotions. Also, a strict threshold would create a sparse dataset, limiting training as good data is already scarce. So, this module works with the uncertainty modeling module, with it enhancing input quality first and the latter handling remaining uncertainties for more robust GER.

Concretely, we adopt a SER-FIQ face quality assessment~\cite{terhorst2020ser}. This methodology efficiently filters out nearly unrecognizable facial samples before passing them to the feature extractor, thus enhancing the quality of the input samples in a unique manner. This ensures that only data of the highest possible reliability and quality is utilized, thereby enhancing the robustness and performance of the model.

Formally, given a face sample $I$, the face quality score $s(I)$ can be obtained by the pre-trained face recognition model. Let $F_{raw}=\{I_{N}\}$ denote $N$ raw facial samples directly detected from the image.
The quality scores $S=\{s_{1},s_{2},\cdots,s_{N}\}$ of $F_{raw}$ are acquired by using face quality assessment strategy in  \cite{terhorst2020ser}. Concretely, the quality of an image is estimated by calculating the pairwise distances between different stochastic embedding, which is obtained through different random sub-networks of face recognition. The face quality score is formulated as,
\begin{eqnarray}
S_{k}(X(I_{k})) &=& 2 \sigma(-\frac{2}{m^2}\sum_{i<j}d(x_{k_{i}},x_{k_{j}})),
\end{eqnarray}
where $S_{k}$ is the quality score corresponding to the $k$-th face image. 
$X(\cdot)$ is a set with $m$ face embeddings acquired from different face recognition model. $d(x_{k_{i}}, x_{k_{j}})$ means the Euclidean distance between the randomly selected embeddings pairs $x_{k_{i}}$ and $x_{k_{j}}$.  

Especially, the image enhancement module equals face image quality estimation (FIQE) to fetch the samples with the high score by setting a threshold. The filtered $F_{input}$ can be defined as,
\begin{eqnarray}
F_{input} = \{I_{i}| s_{i}\ge\delta_{2},  \forall i \in R^{N} \},
\label{eqn:FIQE}
\end{eqnarray}
where $F_{input}$ is the facial set as the final input of the face branch, which the severely low-quality samples have been discarded, $\delta_{2}$ is a pre-defined threshold. 

\subsection{Proportional-weighted Fusion Strategy}
The previous methods \cite{guo2020graph,fujii2020hierarchical,KhanLCMOT18} on the GER task demonstrated the improvements in performance that can be obtained by fusing multiple emotion cues of different compositions which contain complementary information. Similarly, we incorporate three branches (\ie, face, object, and scene) into the proposed framework, which from the global and local perspectives acquire group-level emotion features. Given the group-level features extracted from each branch (\ie, $x^{f}_{group}$, $x^{o}_{group}$, $x^{s}_{group}$), the prediction scores for GER via the classifier are obtained, $sc^{f} = C^{f}(x^{f}_{group}), sc^{o} = C^{o}(x^{o}_{group}), sc^{s} = C^{s}(x^{s}_{group})$, where $sc^{f}$, $sc^{o}$, and $sc^{s}$ are prediction scores for face, object, and scene branch, respectively. $C^{f}(\cdot)$, $C^{o}(\cdot)$, and $C^{s}(\cdot)$ are the corresponding classifiers.

The extensive score-level fusion strategy mainly focuses on two forms. (1) The first is to fuse scores by using arithmetic-based measures, \ie, a weighted average. (2) The second is to employ a grid search approach to find the optimal fusion weights. Crucially, both strategies are required to learn fusion weights empirically from the validation set and fixed when testing. However, the label of the test set for whole GAFF databases in our experiment is available only for those participating in the EmotiW competitions, we cannot but obey the official protocol to use the validation set to verify the performance of our model. Hence, neither of the above-mentioned two fusion strategies is usable for our method.

In our problem, we design a proportional-weighted fusion strategy (PWFS) to account for the corresponding proportion as a weight for each branch which is able to adequately utilize
the potential and complementary information. Specifically,  we approximately calculate the final group-level emotion prediction score as:
\begin{eqnarray}
sc^{group} = w^{f}sc^{f} + w^{o}sc^{o} + w^{s}sc^{s},
\end{eqnarray}
where $w^{f}$, $w^{o}$, and $w^{s}$ are the proportion of each corresponding branch prediction score to the sum of the whole branches. And $sc^{group}$ represents the final classification score, in which every branch score is integrated by the proportional weight. Furthermore, due to the small size of the GER dataset, the proportional-weighted fusion separately processes each branch for GER which can avoid causing the fusion results to be worse than a single branch attributed to over-fitting. To reason more independently produces better results.
\section{Experimental and Discussion}
\label{sec:experiment}

\begin{table}
\centering
\small
\caption{\textcolor{black}{Statistics of the two GAFF databases.}}
\begin{tabular}{c|c|ccc|c}		
\hline
Dataset & Type & Positive & Neutral & Negative & Total\\
\hline
\multirow{2}*{GAFF2} & train & 1272 & 1199 & 1159 & 3630\\
~ & val & 773 & 728 & 564 & 2065 \\
%~ & test & - & - & - & -  \\
\hline
\multirow{2}*{GAFF3} & train & 3977 & 3080 & 2758 & 9815\\
~ & val & 1747 & 1368 & 1231 & 4346 \\
%~ & test & - & - & - & - \\
%~ & test & 829 & 916 & 1266 & 3011
\hline
\end{tabular}
\label{tab:database}
\end{table}

\subsection{Databases and Evaluation Metrics}
\noindent\textbf{GAFF Databases.} The GAFF databases consist of two series benchmark databases: the Group AFFective 2.0 (GAFF2) \cite{dhall2017individual} database and the Group AFFective 3.0 (GAFF3) \cite{dhall2018emotiw} database. All samples in GAFF databases are collected from the Internet by searching for keywords such as protest, violence, festival, etc, and each sample contains at least two people. The statistics of the two databases are summarized in Table~\ref{tab:database}. The total number of images for each category is listed in this table. All the samples are annotated with three emotion categories: positive, negative, and neutral. As the label is not released in the test set, and while only available to those participating in the EmotiW competitions \cite{dhall2017individual, dhall2018emotiw}. We only conduct all experiments on the training and validation sets. More specifically, we train our model on the training set and use the validation set to verify the performance of our model.

\noindent\textbf{MultiEmoVA Database.} The MultiEmoVA database \cite{MouCG15} was collected by using keywords such as graduation, party, etc, from Google Images and Twitter. It was fused by arousal-level and valence-level to annotate six categories as high-positive, medium-positive, high-negative, medium-negative, low-negative, and neutral, which the corresponding number of samples is 46, 64, 31, 27, 10, and 72 images, respectively. 

\noindent\textbf{Evaluation Metrics.} To evaluate our method, we utilize three performance metrics on the GAFF databases, which are Recall rate, including normal average and unweighted average recall (UAR), Precision rate, and F-measure following previous methods \cite{fujii2019hierarchical,fujii2020hierarchical,abbas2017group,surace2017emotion}. Following the experiment setup in \cite{huang2019analyzing}, we formulate the experiments on the MultiEmoVA database as a 5-class classification task  (\ie, medium-negative, high-negative, medium-positive, high-positive, and neutral). And we use 5-fold-cross-validation
protocol and reported the average recognition accuracy.\

\begin{table*}[]
% \begin{sidewaystable} 
\small
\centering
\caption{\textcolor{black}{Performance Comparison with the state-of-the-art methods on the GAFF2 database. The best results are in bold, and the underline means second better.}}
\label{Tab03}
\begin{tabular}{l|ccc|cc|ccc|c|ccc|c}
\hline
\multirow{2}{*}{Methods} & \multicolumn{5}{c|}{Recall} & \multicolumn{4}{c|}{Precision} & \multicolumn{4}{c}{F-measure} \\

\cline{2-6} \cline{7-10} \cline{11-14}
& Pos. & Neu. & Neg. & Ave. & UAR
& Pos. & Neu. & Neg. & Ave.
& Pos. & Neu. & Neg. & Ave.  \\
\cline{1-14}
Dhall $et$ $al.$ \cite{dhall2017individual} &- & - & - & - & 52.97 & - & - & - & - & - & - & - & - \\

Shamsi $et$ $al.$ \cite{abs-1710-01216} &- & - & - & - & 55.23 & - & - & - & - & - & - & - & -  \\

Sokolov $et$ $al.$ \cite{SokolovP18} &- & - & - & - & 64.89 & - & - & - & - & - & - & - & -  \\

Surace $et$ $al.$ \cite{surace2017emotion} &79.43 & 61.26& 65.33 & 68.68 & 67.75 & 68.61 & 59.63 & 76.05 & 67.75 & 73.62 & 60.43 & 70.29 & 68.11 \\

Bawa $et$ $al.$ \cite{bawa2019emotional} &- & - & - & - & 68.53 & - & - & - & - & - & - & - & -   \\

Balaji $et$ $al.$ \cite{BalajiO17} &- & - & - & - & 71.50 & - & - & - & - & - & - & - & -   \\

Huang $et$ $al.$ \cite{huang2019analyzing} &- & - & - & - & 72.17 & - & - & - & - & - & - & - & -   \\

Abbas $et$ $al.$ \cite{abbas2017group} &78.53 & 65.28& 72.70 & 72.17 & 72.38 & 79.76 & 66.20 & 69.97 & 71.98 & 79.14 & 65.74 & 71.30 & 72.06 \\

Fujii $et$ $al.$ \cite{fujii2019hierarchical} &\uline{86.93} & 67.45& 64.73 & 73.30 & 74.00 & 75.68 & 69.64 & \textbf{77.33} & 74.22 & 80.92 & 68.53 & 70.46 & 73.30 \\

Rassadin $et$ $al.$ \cite{RassadinGS17} &80.00 & 66.00 & \textbf{80.00} & 75.33 & 75.39 & - & - & - & - & - & - & - & -   \\

Tarasov $et$ $al.$ \cite{TarasovS18} &80.00 & 72.00 & 74.00 & 75.33 & 75.50 & - & - & - & - & - & - & - & -   \\

Wei $et$ $al.$ \cite{WeiZXL0YS17} &- & - & - & - & 77.92 & - & - & - & - & - & - & - & -   \\

Zhang $et$ $al.$ \cite{zhang2022semi} &- & - & - & - & 78.51 & 85.38 & \textbf{84.49} & 60.89 & 76.92 & - & - & - & -   \\

Fujii  $et$ $al.$ \cite{fujii2020hierarchical} &\textbf{88.41} & \uline{72.51} & \uline{79.64} & \textbf{80.19} & \textbf{80.41} & \textbf{87.84} & \uline{77.55} & 74.10 & \textbf{80.19} & \textbf{88.12} & \textbf{74.95} & 76.76 & \textbf{79.95} \\

Ours & 84.16 & \textbf{75.18} & 78.62 & \uline{79.32} & \uline{79.52} & \uline{87.57} & 73.93 & \uline{76.08} & \uline{79.19} & \uline{85.83} & \uline{74.55} & \textbf{77.33} & \uline{79.23} \\
\hline
\end{tabular}
\label{tab:GAFF2}
\end{table*}
% \end{sidewaystable}

\begin{table*}[]
\small
\centering
\caption{Comparison with the state-of-the-art methods on the GAFF3 database. The best results are in bold, and the underline means second better.}
\label{Tab03}
\begin{tabular}{l|ccc|cc|ccc|c|ccc|c}
\hline
\multirow{2}{*}{Methods} & \multicolumn{5}{c|}{Recall} & \multicolumn{4}{c|}{Precision} & \multicolumn{4}{c}{F-measure} \\

\cline{2-6} \cline{7-10} \cline{11-14}
& Pos. & Neu. & Neg. & Ave. & UAR
& Pos. & Neu. & Neg. & Ave.
& Pos. & Neu. & Neg. & Ave.  \\
\cline{1-14}
Dhall $et$ $al.$ \cite{dhall2018emotiw} &- & - & - & - & 65.00 & - & - & - & - & - & - & - & - \\

Garg $et$ $al.$ \cite{Garg2019} &- & - & - & - & 65.27 & - & - & - & - & - & - & - & -  \\

Nagarajan $et$ $al.$ \cite{NagarajanO19} &- & - & - & - & 70.10 & - & - & - & - & - & - & - & -  \\

Fujii $et$ $al.$ \cite{fujii2019hierarchical} &\uline{88.31} & 60.40 & 58.85 & 69.19 & 71.27 & 72.12 & 69.51 & 71.52 & 71.05 & 79.40 & 64.64 & 64.57 & 69.53  \\

Gupta $et$ $al.$ \cite{GuptaACDP18} &- & - & - & - & 74.38 & - & - & - & - & - & - & - & -   \\

Quach $et$ $al.$ \cite{quach2022non} &85.00 & \textbf{84.00} & 53.00 & 74.00 & 76.12 & - & -  & - & 74.18 & - & - & - & 73.81  \\

Dejian $et$ $al.$ \cite{Dejian2020Group} &- & - & - & - & 76.30 & - & - & - & - & - & - & - & -   \\

Zhang $et$ $al.$ \cite{zhang2022semi}  &- & - & - & - & 77.01 & 79.85 & \textbf{76.61} & 73.44 & \uline{76.63} & - & - & - & -   \\

Fujii $et$ $al.$ \cite{fujii2020hierarchical} &\textbf{89.83} & \uline{70.92} & \uline{67.41} & \uline{76.05} & \uline{77.54} & \uline{82.88} & \uline{72.64} & \textbf{74.32} & 76.61 & \textbf{86.21} & \textbf{71.77} & \uline{70.69} & \uline{76.23}  \\

Ours &87.12 & 69.81 & \textbf{74.09} & \textbf{77.01} & \textbf{77.98} & \textbf{84.46} & 72.62 & \uline{74.21} & \textbf{77.10} & \uline{85.77} & \uline{71.19} & \textbf{74.15} & \textbf{77.04} \\
\hline
\end{tabular}
\label{tab:GAFF3}
\end{table*}

% \begin{table*}[]
% \small
% \centering
% \caption{Comparison of the complexity comparisons with methods that exhibit relatively superior classification performance. "# Param." indicates the total number of the training model parameters.}
% \label{Tab03}
% \begin{tabular}{l|cc|cc|c}
% \toprule
% \multirow{2}{*}{Methods} & \multicolumn{2}{c|}{Feature Extraction} & \multicolumn{2}{c|}{Feature Aggregation} & \multirow{2}{*}{#Param.} \\

% \cline{2-3} \cline{4-5} \
% & Type & Num. & Type & Num. \\
% \cline{1-6}
% Fujii $et$ $al.$ \cite{fujii2020hierarchical} & VGG16 & 3 & Attention & 2 & 356M  \\

% Ours  & ResNet18 \& VGG19 & 3 & Uncertainty score & 1 & 282M\\

% \bottomrule
% \end{tabular}
% \label{tab:Complex}
% \end{table*}

\subsection{Implementation Details}
As shown in the pipeline, the proposed method is a three-branch framework, which consists of a face branch, an object branch, and a scene branch. Each branch is trained independently for GER. In our training process, the images of each branch are performed the standard transformations for data augmentation, which are resizing, random rotation, random horizontal flipping, and normalization. For the face branch, the face samples in all databases are resized to $224\times224$ pixels. Similar with \cite{ChangLCW20}, the uncertainty encoders $f_{\theta_{\mu}}$ and $f_{\theta_{\sigma}}$ are implemented by BackBone-Flatten-FC-BN to output 512-dimension feature embeddings in the face branch. We employ Adam as the optimizer with an initial learning rate of 0.0001. For the branch and scene branches, we resize the input samples to $256\times 256$ pixels and the optimizer is stochastic gradient descent (SGD) with a learning rate 0.0001. The uncertainty encoders in the object branch are two FC layers and the $\lambda_{1}$ (\cite{ChenLD0L22}) in the classification loss term is set to 0.1. For the hyper-parameters in the UAL module, the weight of the KL regularization term is $1e^{-4}$ for $L_{kl}(\lambda_{2})$ (\cite{ChenLD0L22}) and the rank regularization term $L_{rank}(\lambda_{3})$ (\cite{WangPYL020}) is equal to the weight of $L_{cls}$ in Eq.~\ref{eqn:final_face_loss}. The weight of $L_{rec}$ is set to $0.01 (\lambda_{4})$ (\cite{ChenLD0L22}). The $\beta$ and $\delta_{1}$ in the rank regularization are set as 0.5 and 0.2, respectively. The $\delta_{2}$ in the FIQE module is set as 0.3. Training batch size and epoch are set as 64 and 100, respectively. The setting is the same on all databases except for the two more data augmentations (\ie,  colorJitter and random vertical flip) on the MultiEmoVA database. The proposed method is implemented using PyTorch and trained on an RTX 3090 GPU.

\subsection{Comparison with the State-of-the-Art (SOTA) methods}
We compared the proposed method with state-of-the-art methods that aim to improve GER performance by exploring diverse emotion-related cues from multiple sources and subsequently aggregating individual features into group-level insights~\cite{huang2019analyzing,fujii2019hierarchical,quach2022non,fujii2020hierarchical,abs-1710-01216,TarasovS18,NagarajanO19,surace2017emotion,BalajiO17,abbas2017group,Garg2019,GuptaACDP18,RassadinGS17,Dejian2020Group,bawa2019emotional,WeiZXL0YS17,zhang2022semi,dhall2017individual,dhall2018emotiw,SokolovP18,MouCG15} on the GAFF2, GAFF3, and MultiEmoVA databases. Additionally, we can roughly categorize different aggregation methods based on individual characteristics into the following types: arithmetic
averaging~\cite{surace2017emotion,BalajiO17,abbas2017group}, voting~\cite{fujii2019hierarchical,Garg2019,GuptaACDP18}, Recurrent Neural Networks (RNNs)~\cite{bawa2019emotional,WeiZXL0YS17,quach2022non}, and
attention mechanisms\cite{fujii2020hierarchical}. It is worth noting that our method shares similar with attention-based approaches, as both aim to learn efficient fusion weights for aggregating individual features. However, our method explicitly models uncertainty among individuals and approximates the uncertainty of each individual as weights for aggregating individuals, enhancing the diversity and robustness of representations to GER.

1) Results on the GAFF2 dataset: The results on the GAFF2 dataset are reported in Table~\ref{tab:GAFF2}. From Table~\ref{tab:GAFF2}, our proposed method outperforms nearly all SOTA approaches, with only slight variations compared to~\cite{fujii2020hierarchical} in certain emotion categories. Specifically, our method exhibits a marginal decrease in performance, with average decreases of 0.87\%, 1\%, and 0.72\% in terms of Recall, Precision, and F-measure, respectively. It is noteworthy that~\cite{fujii2020hierarchical} employs a hierarchical classification approach. Initially, it conducts binary classification to determine whether a sample is "Positive" or "Not Positive" based on facial expression features. Subsequently, for samples categorized as "Not Positive," they further conduct the final three-class emotion recognition task using scene and object emotion features. Therefore, ~\cite{fujii2020hierarchical} achieve superior metrics for the "Positive" category, attributed to its hierarchical classification approach effectively distinguishing "Positive" images. Additionally, their approach sacrifices high computational power to extract more useful information in the object branch, contributing to their superior performance. In contrast, our method not only abstains from using a hierarchical framework but also solely employs a straightforward CNN-based network as the encoder to acquire object information, significantly reducing computational demands. As comparative analysis of method complexity between our approach and ~\cite{fujii2020hierarchical} is conducted in Table~\ref{tab:Complex}. This analysis illustrates that our approach achieves a competitive recognition performance while simultaneously reducing execution time compared to~\cite{fujii2020hierarchical}. Furthermore, our method surpasses several RNNs-based methods, such as ~\cite{bawa2019emotional,WeiZXL0YS17}, indicating that our method's ability to discriminatively learn the importance of different individuals and utilize this information to aggregate individual features is crucial for achieving high performance in recognition accuracy.

2) Results on the GAFF3 dataset: We further provide detailed comparisons with previous methods listed in Table~\ref{tab:GAFF3} on the GAFF3 dataset. From Table~\ref{tab:GAFF3}, our method achieves state-of-the-art-comparable performance. In particular, we surpass the current best method, the hierarchical framework proposed by~\cite{fujii2020hierarchical}, with improvements of 0.96\% in terms of Recall, 0.49\% in terms of Precision, and 0.81\% in terms of F-measure. This advancement can be attributed to our approach of modeling latent uncertainty across all individuals, specifically formulating uncertainty-sensitive score allocations to facilitate the aggregation of individual features, rather than relying excessively on the utilization of attention modules as in approach~\cite{fujii2020hierarchical}. This allows us to better enhance the robustness of individual and group representation synchronously.

\begin{figure}[t!]
\includegraphics[width=\linewidth]{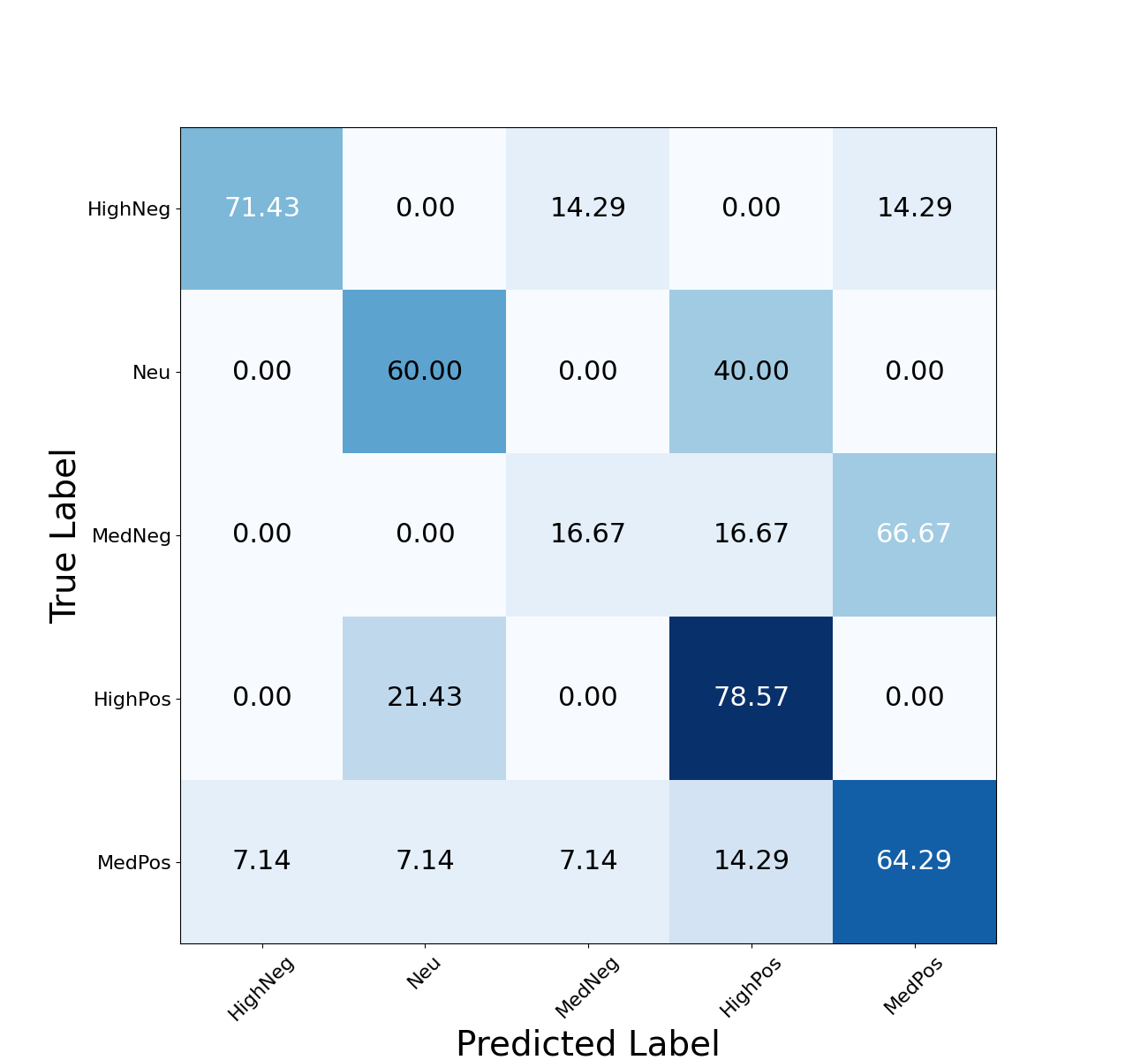}
\caption{Confusion matrices on the MultiEmoVA dataset. The values reflect classification accuracy for every category. “HighPos”, "MedPos"," HighNeg", "MedNeg", and “Neu” are abbreviations for “High-Positive”, "Medium-Positive", "High-Negative", "Medium-Negative", and “Neutral” in the group emotion labels for the MultiEmoVA dataset, respectively. }
\label{fig:confusion matrices}
\end{figure}

3) Results on the MultiEmoVA dataset: The aggregate sample volume within the MultiEmoVA dataset significantly lags behind that of the GAFF databases. Despite the reduced number of training samples, Table~\ref{tab:MultiEmoVA} demonstrates our method outperforms all state-of-the-art methods with a considerable margin for group-level emotion recognition accuracy. Compared to the method by Mou~\etal~\cite{MouCG15}, which exhibits the lowest performance, our method achieves a remarkable 21.26\% improvement in UAR. Mou~\etal~\cite{MouCG15} introduced the MultiEmoVA dataset and provided a baseline algorithm, which directly concatenates multiple representations from face, body, and scene into one representation vector for final emotion classification. Additionally, the method by Huang~\etal~\cite{huang2019analyzing}, which emphasizes global alignment kernels and formulates the Support Vector Machine with combined global alignment kernels, aims to enhance the recognition of group-level emotion and currently achieves the highest accuracy. However, our method surpasses it by 6.82\%. To provide deeper insights into the classification results, we visualize the confusion matrices on the MultiEmoVA dataset. Fig.~\ref{fig:confusion matrices} illustrates the visualization results. According to this figure, four included group emotions, \ie, high-positive, medium-positive, high-negative, and neutral, are easier to distinguish compared to medium-negative. We attribute this observation to class imbalance, as well as tendency for medium-negative samples to be classified as medium-positive.

Meanwhile, our method achieves a performance of 60.77\% in terms of the F-measure metric. It supports our conclusion that our method enhances feature discrimination ability and improves the model's generalization, even under conditions of extreme sample scarcity and severe class imbalance, while still achieving good performance. In summary, the results across multiple datasets demonstrate the effectiveness of our proposed method and the feasibility of mitigating uncertainty in group-level emotion recognition.

4) Results on the GAFF2 Contracted Dataset: We built a sub-database comprising challenging scenarios, such as occlusions, to evaluate our method and other methods on the GER task. For the construction of the sub-database, we manually selected approximately 40\% of the samples from each class of the validation set of the GAFF2 dataset, focusing on instances where individual occlusions or occlusions between individuals were present. Regarding the selected comparative methods, as there are currently no publicly accessible codes for these methods, we replicated experiments comparing our method with two benchmark models (\ie, VGG16 and ResNet50) that performed well on the GER task. The experimental results are shown in Table~\ref{tab:subGAFF2}. Our method exhibits a minor decrease in performance under the prevalent occlusions in the group and maintains superior performance.

\begin{table}
\centering
\small
\caption{\textcolor{black}{Comparison with the state-of-the-art methods on the MultiEmoVA database.}}
\begin{tabular}{l|c|c}		
\hline
Methods & UAR & F-measure\\
\hline
Mou $et$ $al.$ \cite{MouCG15} & 39.96 & -\\

Huang $et$ $al.$ \cite{huang2019analyzing} & 54.40 & -\\

Ours & \textbf{61.22} & 60.77\\
\hline
\end{tabular}
\label{tab:MultiEmoVA}
\end{table}

%%% Model Complex Table 
\begin{table}[]
\small
\centering
\caption{Comparison of the complexity with methods that exhibit relatively superior classification performance. “Cls” indicates the classification methods, and “\# Param.” indicates the total number of the training model parameters.}
\label{Tab04}
\begin{tabular}{l|c|c|c}
\hline
Method & Feature Aggregation & Cls & \# Param.\\
\hline
Fujii $et$ $al.$  \cite{fujii2020hierarchical} & Cascade Attention  & Binary & 356M  \\

Ours  & Uncertainty Score & Primary & 282M\\
\hline
\end{tabular}
\label{tab:Complex}
\end{table}

\begin{table*}[]
\small
\centering
\caption{\textcolor{black}{Comparison with the state-of-the-art methods on the different constructed GAFF2 database. “w/o cut” indicates the intact GAFF2 database and “w/ cut” indicates the truncated sub-dataset of GAFF2. “*” represents the results we reproduced.}}
\label{Tab03}
\begin{tabular}{l|l|ccc|cc|ccc|c|ccc|c}
\hline
\multirow{2}{*}{Dataset} & \multirow{2}{*}{Methods} & \multicolumn{5}{c|}{Recall} & \multicolumn{4}{c|}{Precision} & \multicolumn{4}{c}{F-measure} \\

\cline{3-7} \cline{8-11} \cline{12-15}
& & Pos. & Neu. & Neg. & Ave. & UAR
& Pos. & Neu. & Neg. & Ave.
& Pos. & Neu. & Neg. & Ave.  \\
\cline{1-15}

 \multirow{5}{*}{w/o cut} & ResNet50  &76.46 & 61.21 & \textbf{82.29} & 73.31 & 72.68 & - & - & - & - & - & - & - & - \\~
%\cite{zhang2022semi}
 & ResNet50*  &76.84 & 67.70 & 75.88 & 73.48 & 73.38 & 87.35 & 64.95 & 67.94 & 73.42 & 81.76 & 66.30 & 71.69 & 73.25 \\~

 & VGG16  &78.53 & 65.28& 72.70 & 72.17 & 72.38 & 79.76 & 66.20 & 69.97 & 71.98 & 79.14 & 65.74 & 71.30 & 72.06 \\~
%\cite{abbas2017group}
 & VGG16*  &78.14 & 62.91 & 78.85 & 73.30 & 72.98 & 83.89 & 67.17 & 66.72 & 72.59 & 80.91 & 64.97 & 72.28 & 72.72  \\~

 & Ours & \textbf{84.16} & \textbf{75.18} & 78.62  & \textbf{79.32} & \textbf{79.52} & \textbf{87.57} & \textbf{73.93} & \textbf{76.08} & \textbf{79.19} & \textbf{85.83} & \textbf{74.55} & \textbf{77.33} & \textbf{79.23} \\

\hline

 \multirow{3}{*}{w/ cut} & ResNet50* &64.55 & 62.12& 78.18 & 68.28 & 67.36 & 86.94& 57.59 & 62.77 &69.10 & 74.09 & 59.77 & 69.64 & 67.83 \\~

 & VGG16* &58.19 & 55.30& \textbf{85.91} & 66.46 & 64.66 & 86.57 & 60.00 & 55.43 & 67.33 & 69.60 & 57.55 & 67.38 & 64.84 \\~

 & Ours &\textbf{75.92} & \textbf{65.19} & 83.18 & \textbf{74.76} & \textbf{74.01} & \textbf{89.02} & \textbf{68.71} & \textbf{65.59} & \textbf{74.44} & \textbf{81.95} & \textbf{66.90} & \textbf{73.35} & \textbf{74.07} \\
\hline
\end{tabular}
\label{tab:subGAFF2}
\end{table*}

\begin{table*}[]
\small
\centering
\caption{\textcolor{black}{Effect on different loss terms on the GAFF2 database.}}
\label{Tab03}
\begin{tabular}{p{0.1cm}p{0.2cm}p{0.2cm}p{0.5cm}p{0.4cm}|ccc|cc|ccc|c|ccc|c}
\hline
\multicolumn{5}{c}{Loss terms} & \multicolumn{5}{|c}{Recall} & \multicolumn{4}{|c}{Precision} & \multicolumn{4}{|c}{F-measure}\\
\hline
No. & $L_{cls}$ & $L_{kl}$ & $L_{rank}$ & $L_{rec}$ & Pos. & Neu. & Neg. & Ave. & UAR
& Pos. & Neu. & Neg. & Ave. & Pos. & Neu. & Neg. & Ave.  \\
\hline
1 & \Checkmark & \XSolidBrush & \XSolidBrush & \XSolidBrush & 76.36 & 55.43 & \textbf{83.83} & 71.87 & 71.00 & \textbf{86.73} & 68.95 & 58.65 & 71.44 & 81.22 & 61.45 & 69.01 & 70.57 \\
2 & \Checkmark & \Checkmark& \XSolidBrush & \XSolidBrush  & 81.56 & 60.37 & 77.51 & 73.14 & 73.03  & 84.64 & 68.70 & 63.96 & 72.43  & 83.07 & 62.64 & 70.09 & 72.47 \\
3 & \Checkmark & \Checkmark & \Checkmark & \XSolidBrush  & 81.95 & 64.17 & 74.35 & 73.49 & 73.67  & 84.93 & 68.11 & 66.01 & 73.02  & 83.41 & 66.09 & 69.93 & 73.14 \\
4 & \Checkmark & \Checkmark & \Checkmark &  \Checkmark & \textbf{82.99} & \textbf{65.02} & 76.58 & \textbf{74.86} & \textbf{74.96} & 84.75 & \textbf{70.49} & \textbf{67.65} & \textbf{74.30} & \textbf{83.86} & \textbf{67.64} & \textbf{71.84} & \textbf{74.45}\\

\hline
\end{tabular}
\label{tab:Ablation_GAF2}
\end{table*}

\begin{table*}[]
% \begin{sidewaystable}
\small
\centering
\caption{\textcolor{black}{Ablation results on the GAFF2 database by using different variants.}}
\label{Tab03}
\begin{tabular}{cl|ccc|cc|ccc|c|ccc|c}
\hline
\multirow{2}{*}{No.} & \multirow{2}{*}{Methods} & \multicolumn{5}{c|}{Recall} & \multicolumn{4}{c|}{Precision} & \multicolumn{4}{c}{F-measure} \\

\cline{3-7} \cline{8-11} \cline{12-15}
&& Pos. & Neu. & Neg. & Ave. & UAR
& Pos. & Neu. & Neg. & Ave.
& Pos. & Neu. & Neg. & Ave.  \\
\cline{1-15}
5& \makecell[l]{Face w/o \\UAL\&FIQE}  & 70.26 & 63.33 & 73.79 & 69.13 & 68.77 & 83.49 & 60.03 & 63.93 & 69.15 & 76.30 & 61.63 & 68.51 & 68.82 \\

6& Face w/o UAL & 73.77 & 61.35 & 75.09 & 70.07 & 69.76 & 84.27 & 62.59 & 62.35 & 69.74 & 78.67 & 61.97 & 68.13 & 69.59 \\

7 & Face w/o FIQE  & 82.99 & 65.02 & 76.58 & 74.86 & 74.96 & 84.75 & 70.49 & 67.65 & 74.30 & 83.86 & 67.64 & 71.84 & 74.45 \\

8 & OnlyFace & \textbf{85.71} & 65.73 & 74.72 & 75.39 & 75.76  & 86.16 & 71.25 & 67.34 & 74.92 & \textbf{85.94} & 68.38 & 70.84 & 75.05 \\

\hline

9& Object  w/o UAL  & 76.97 & 72.36 & 63.08 & 70.80 & 71.65 & 80.84 & 62.48 & 73.28 & 72.20 & 78.86 & 67.06 & 67.80 & 71.24 \\

10& OnlyObject & 76.71 & 72.21 & 67.35 & 72.09 & 72.64 & 81.23 & 64.08 & 73.78 & 73.03 & 78.91 & 67.90 & 70.42 & 72.41 \\

\hline

11& OnlyScene & 75.97 & 69.82 & 78.07 & 74.62 & 74.37 & 84.91 & 67.44 & 70.71 & 74.35 & 80.19 & 68.61 & 74.20 & 74.33 \\

12 &\textbf{Ours}  & 84.16 & \textbf{75.18} & \textbf{78.62} & \textbf{79.32} & \textbf{79.52} & \textbf{87.57} & \textbf{73.93} & \textbf{76.08} & \textbf{79.19} & 85.83 & \textbf{74.55} & \textbf{77.33} & \textbf{79.23} \\

\hline
\end{tabular}
\label{tab:Ablation_GAF2_whole}
\end{table*}
% \end{sidewaystable}

\begin{table*}[]
\small
\centering
\caption{Fusion ablation of face, object, and scene branch on the GAFF2 database.}
\label{Tab03}
\begin{tabular}{cl|ccc|cc|ccc|c|ccc|c}
\hline
\multirow{2}{*}{No.} & \multirow{2}{*}{Methods} & \multicolumn{5}{c|}{Recall} & \multicolumn{4}{c|}{Precision} & \multicolumn{4}{c}{F-measure} \\

\cline{3-7} \cline{8-11} \cline{12-15}
& & Pos. & Neu. & Neg. & Ave. & UAR
& Pos. & Neu. & Neg. & Ave.
& Pos. & Neu. & Neg. & Ave.  \\
\cline{1-15}
13& Equal proportion &83.25 & 74.19 & 72.68 & 76.70 & 77.24 & 85.47 & 70.13 & 75.63 & 77.08 & 84.34 & 72.10 & 74.12 & 76.86 \\

14& Global priority & 82.73 & 74.89 & 73.79 & 77.14 & 77.59 & 86.31 & 70.33 & 75.76 & 77.47 & 84.48 & 72.54 & 74.76 & 77.26 \\

15& Face priority & \textbf{84.42} & 73.91 & 74.35 & 77.56 & 78.04 & 86.55 & 71.39 & 75.19 & 77.71 & 85.47 & 72.63 & 74.77 & 77.62   \\

16& \textbf{PWFS (ours)} & 84.16 & \textbf{75.18} & \textbf{78.62} & \textbf{79.32} & \textbf{79.52} & \textbf{87.57} & \textbf{73.93} & \textbf{76.08} & \textbf{79.19} & \textbf{85.83} & \textbf{74.55} & \textbf{77.33} & \textbf{79.23} \\
\hline
\end{tabular}
\label{tab:Ablation_Fusion}
\end{table*}

\subsection{Ablation Study}
We conduct ablation studies (marked by ``No.'') to investigate the contributions of $L_{kl}$, $L_{rank}$, and $L_{rec}$ in Eq.~\ref{eqn:final_face_loss}, the module components and the fusion strategy by removing constituent components on the GAFF2 database, to validate the effectiveness and respective contributions of the model. Furthermore, we also verify the effect of the total sample time  $N$ of the inference stage.

\textbf{Impact of different loss terms.} As shown in Eq.~\ref{eqn:final_face_loss}, 4 loss terms are considered in our proposed method. The $L_{cls}$ represents the baseline cross-entropy loss for general GER methods. Besides, $L_{kl}$, $L_{rank}$, and $L_{rec}$ are proposed for uncertainty-aware learning module. We start with the exploration of the effectiveness of different loss terms. The results are reported in Table~\ref{tab:Ablation_GAF2}.

Compared with sole utilizing $L_{cls}$ as shown in ``No. 1'' of Table~\ref{tab:Ablation_GAF2}, the KL divergence term $L_{kl}$ (in ``No. 2'' of Table~\ref{tab:Ablation_GAF2}) gains the considerable improvements in terms of all metrics. The performance improvement emphasizes the importance of alleviating negative uncertainties. The rank regularization term $L_{rank}$ (in ``No. 3'' of Table~\ref{tab:Ablation_GAF2}) is employed to explicitly constrain the uncertainty-sensitive score of the face individuals, and further regularize the important scale weights which are devised to aggregate the individuals in a group. Compared with ``No. 2'' of Table~\ref{tab:Ablation_GAF2}, it yields considerable improvements of 0.35\%, 0.59\%, and 0.67\% in terms of Recall, Precision, and F-measure, respectively. The reconstruction loss $L_{rec}$ (``No. 4'' of Table~\ref{tab:Ablation_GAF2}) is introduced to drop as much ambiguous information as possible, calculating the L1 distance between the original face individual feature and the sampled stochastic features. Intuitively, the introduction of $L_{rec}$ contributes to better performances and improves the average Recall and UAR significantly. Compared with ``No. 1'' of Table~\ref{tab:Ablation_GAF2},  better results are obtained using the uncertainty-aware loss terms, with increases of 3.96\%, 2.86\%, and 3.88\% in terms of Recall, Precision, and F-measure, respectively. The above results prove the effectiveness of the designed UAL module with the uncertainty-aware loss terms in improving the robust representation ability of individuals, which can further increase the final emotion recognition performance.

\textbf{Impact of different components.} We further conduct the ablation studies on different compositions for GER, which involves two aspects, investigate the effect of appending two different modules (\ie, UAL and FIQE) to the face branch or UAL module to the object branch, and explore the performance of single branch or all branches are attached to the overall framework. Note that FIQE is the simple expression for the method used in the image enhancement module. For variants of FIQE and UAL in the face branch, (1) w/o UAL \& FIQE: only a standard baseline ResNet18 network as the encoder, which directly extracts the deterministic embedding and sums up to the group-level features; (2) w/o FIQE: unlike the baseline, this variant helps model uncertainties by sampling operation and outputting the $\mu$ and $\sigma$ to obtain the stochastic representation of each face individuals; (3) w/o UAL: individual samples have to go through FIQE module before they are fed into the feature extractor, to do the enhancement of the training samples; (4) OnlyFace: with the addition of FIQE and UAL modules to the face branch simultaneously, the more robust group-level representations aggregated by face individuals and only the face branch is used for group-level emotion inference. For the variant of UAL in the object branch, (5) w/o UAL: the detected object individuals are straightly fed into the pre-trained VGG network to capture the deterministic embedding for classification with the cross-entropy loss; (6) OnlyObject: the deterministic representation of each object individual is extended to a probabilistic distribution via the UAL and then aggregated to represent the group-level emotion in this experiment with the single object branch. (7) OnlyScene: this variant is to directly learn the global emotion information from the whole image which just depends on the scene branch. (8) Ours: this variant contains three branches used in our method, which handle each branch separately for GER and fuse them by the proposed PWFS.

As shown in Table~\ref{tab:Ablation_GAF2_whole}, the face branch makes remarkable performances integrally boost by 6.0\% compared to the baseline on all metrics. The first three lines of the table (from ``No. 5'' to ``No. 7'' in Table~\ref{tab:Ablation_GAF2_whole}) indicate that appending the FIQE and UAL modules to the baseline improves performance from 68.77\% to 69.76\% and 74.96\% on the metric of UAR, respectively. Especially, the Face w/o FIQE (in ``No. 7'' of Table~\ref{tab:Ablation_GAF2_whole}) improves substantially overall on each metric. These reflect that explicitly modeling the uncertainties brings significant performance improvement. Moreover, combining the above two modules (in ``No. 8'' of Table~\ref{tab:Ablation_GAF2_whole}) can steadily improve the performance of the face branch. 
In addition, there is around 1.0\% performance gain (in ``No. 9'' vs in ``No. 10'' of Table~\ref{tab:Ablation_GAF2_whole}) when attaching UAL to the object branch on all metrics, which further indicates
the effectiveness of the proposed UAL module for the GER task. The result of the scene branch (in ``No. 9'' of Table~\ref{tab:Ablation_GAF2_whole}) shows a performance not much different from the complete face branch. Obviously, the face and scene branches together occupy an important position in the GER task. The last line is the final GER result in our method, which aggregates all information from three branches, resulting in the performance raising dramatically. It indicates that all the ingredients reinforce each other and the combination is important to get remarkable final results.

\textbf{Impact of different fusion.} We also conduct an ablation study on four fusion strategies to combine the face, object, and scene branches in our model. For equal proportion fusion, the weight of each branch is equal, we straightforwardly add all the predictions. Due to the relatively good performance in the face and scene branches (shown in Table~\ref{tab:Ablation_GAF2_whole}), we adopted the strategy of choosing one of the two branches as the priority, respectively. For global priority fusion, we set the weight for the scene branch (global information) to be twice as large as the weight for the sum of the object and face branches (local information), which assumes that the global information contains more information about the group-level emotion-related pieces of information. For face priority fusion, we set the weight for the face branch to be twice as the object and scene branches, which assumes that the face is the most representative carrier of emotion in an image. The results are reported in Table~\ref{tab:Ablation_Fusion}.

According to Table~\ref{tab:Ablation_Fusion}, our proposed PWFS gains the best results of 79.32\%, 79.19\%, and 79.23\% in terms of Recall, Precision, and F-measure, respectively. The equal proportion fusion strategy obtained the results of 76.70\%, 77.08\%, and 76.86\% (in ``No. 13'' of Table~\ref{tab:Ablation_Fusion}) in terms of Recall, Precision, and F-measure, respectively. It indicates that the equal proportion fusion strategy is detrimental to our model, slightly better than using a single face branch. Compared with the global priority fusion strategy, the face priority fusion strategy achieves better performance with an increase of 0.42\% (in ``No. 14'' vs ``No. 15''of Table~\ref{tab:Ablation_Fusion}) in terms of Recall, demonstrating the leading role of the face branch in GER. The experiment results also show that the different branches~\eg, face, and object, cannot be equally treated. Compared with the first three fusion strategies, our PWFS accounts for the corresponding proportion as a weight for each branch, which helps the model to maximally explore the benefit of each branch.

\begin{figure}[t!]
\includegraphics[width=\linewidth]{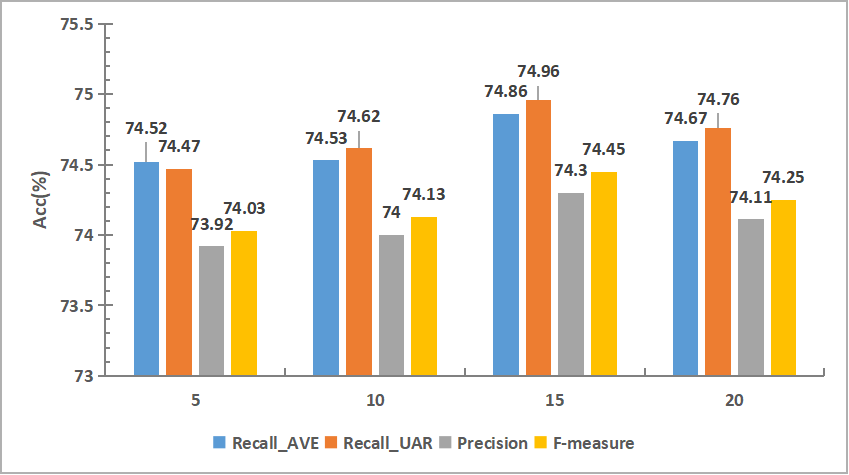}
\caption{\color{black}{Impact of total sample time $M$ on GAFF2 database.}}
\label{fig:sample_time}
\end{figure}

\textbf{Impact of different Monte Carlo sampling time.} To study the effectiveness of sample time $M$ mentioned in  Eq.~\ref{eqn:parameterization}, four different values are chosen to show how $M$ affects the performance on the face branch. Fig.~\ref{fig:sample_time} shows the results on the GAFF2 database. From Fig.~\ref{fig:sample_time}, when $M$ is too small (\eg, $M=5$ or 10), it leads to diminished performance. This is because the limited number of samples may not sufficiently capture the variability within the feature space. Conversely, setting $M$ to a large value,~\eg, $M=20$, may better capture the variability, but the increased sampling would result in higher computational costs and potentially diminish performance. The most suitable value for $M$ is 15, which yields benefits to recognition performance. We hypothesize that this optimal equilibrium allows for effective representation within the feature space while mitigating the influence of randomness, ultimately resulting in improved recognition performance. In all of our experiments, we set $M=15$ as the Monte Carlo sampling time unless specifically defined. Furthermore, as $M$ changes, the recognition performance shows minimal fluctuation, suggesting the robustness and derivability of our model in handling uncertainty.

%  because the appropriate sampling times may arise from achieving an optimal equilibrium between accuracy and computational efficiency. This optimal equilibrium allows for effective representation within the feature space while mitigating the influence of randomness, ultimately resulting in improved recognition performance. 

\begin{figure}[t!]
\includegraphics[width=\linewidth]{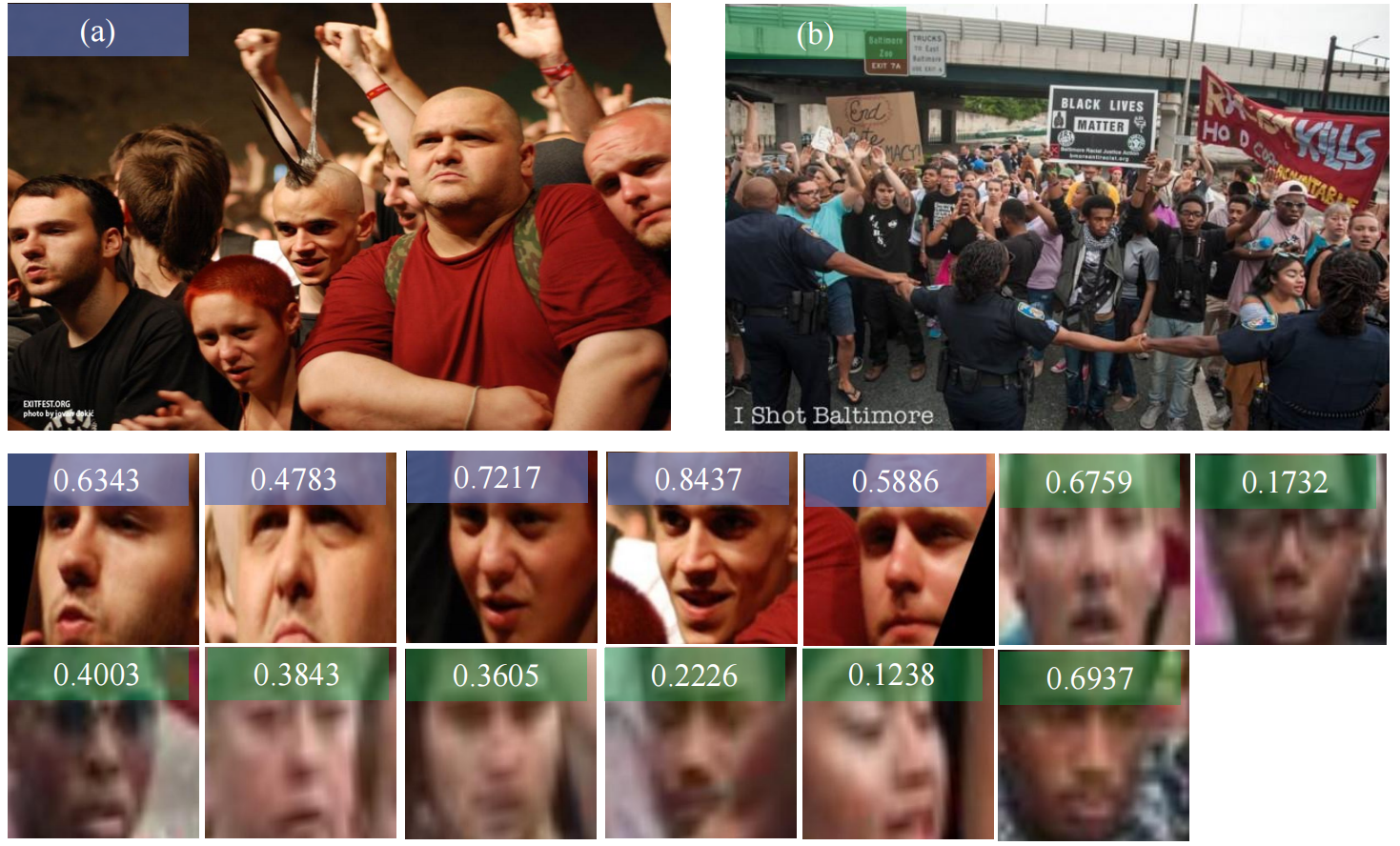}
\caption{\color{black}{Illustration of face quality scores for face individuals within a group from image enhancement module. The threshold $\delta_{2}$ in Eq.~\ref{eqn:FIQE} we set is 0.3, meaning that face individual samples with scores less than 0.3 are discarded in the operations of the image enhancement module.}}
\label{fig:cases_FIQE}
\end{figure}

\begin{figure}[t!]
\includegraphics[width=\linewidth]{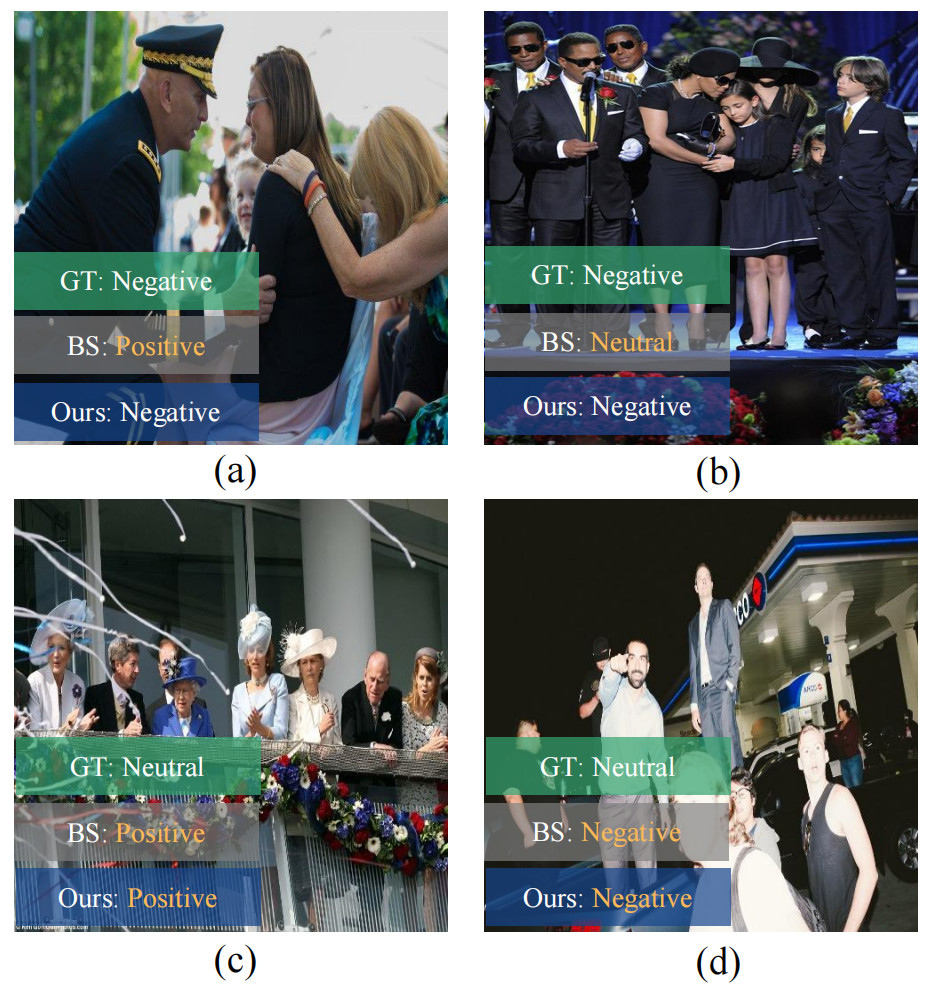}
\caption{\color{black}{Visualization of the group emotion recognition results. ``GT'', ``BS'', and ``Ours'' represent the ground truth, the baseline without uncertainty modeling, and our method. Examples (a) and (b) represent correctly classified results, while (c) and (d) represent misclassified results.}}
\label{fig:cases_model}
\end{figure}

\section{Visualization Analysis}
\label{sec:visualization analysis}
\textbf{Visualization of image enhancement samples.}
To better explore the image enhancement module in our method, we illustrate the face quality scores for two examples of the GAFF2 dataset in Fig.~\ref{fig:cases_FIQE}. In example (a), the sample quality scores for each face individual are generally high, indicating successful passage through the filtering process into the backbone network. Conversely, in example (b), the overall quality scores are lower, suggesting inadequate quality at different levels due to uncertainty. Consequently, discarding a significant number of samples could potentially affect the performance of the facial branch to some extent. To tackle this issue, we will explore methods to enhance resolution directly from the sample source.

\textbf{Visualization of correctly classified and misclassified results.}
Fig.~\ref{fig:cases_model} illustrates the classification results of four samples from the GAFF2 dataset comparing the baseline with our method. In examples (a) and (b), our predictions align with the ground truth, whereas the baseline contradicts it. Our uncertainty modeling, through adaptively assigning fusion weights based on uncertainty-sensitive scores enables robust learning of group-level emotion representations, increasing alignment between ground truth. Furthermore, in examples (c) and (d), show misclassification, where the ground truth ``Neutral'' is predicted as ``Positive'' and  ``Negative'' respectively, by both in the baseline and our method. This could be attributed to the mutual reinforcement of semantic information between dominant individuals and the scene, leading to the model's misjudgment. Alternatively, there exists a potential hypothesis: the subjectivity inherent in label annotations may lead to ambiguity in the ground truth itself. To address this problem, we will explore improvements in the direction of sample label correction and cleaning.

\section{Conclusion}
\label{sec:conclusion}

The lack of research on uncertainty approximation within the realm of GER has been a driving force behind our work. The role of uncertainty approximation is of utmost importance in extending the applicability of emotionally intelligent AI agents to contexts that demand high dependability. This paper introduces our approach, an uncertainty-aware learning method, which seeks to encode latent uncertainty across all individuals, encompassing both faces and objects, while also incorporating scene information for group-level emotion recognition. Our unique contribution lies in explicitly modeling the uncertainty of individual samples as Gaussian random variables, leading to the generation of diverse samples and predictions. We have formulated uncertainty-sensitive score allocations to facilitate the aggregation of individual facial features, thereby yielding more robust GER representations. By employing a sampling operation, we ensure the derivability of the module, while a series of constraints are introduced to mitigate the adverse impact of uncertainty. An image enhancement module has been developed to counteract severe noise in each face individual sample. Additionally, we've designed a PWFS to effectively combine the outputs of three branches, enhancing group-level emotion predictions in GER. Extensive experimentation across three benchmarks validates the efficacy of our approach in managing uncertainty and advancing GER performance.

In summary, our proposed UAL method shows promising success in robust group-level emotion recognition, particularly in congested scenes and noisy environments, but still has limitations in generalizing to extremely complex scenarios with high levels of occlusion, congestion, or other challenging environmental factors not adequately represented in the training data. Further research could focus on refining the model to better address these challenges and improve its overall performance.

It is worth noting that uncertainty estimation within GER still holds substantial potential for improvement. We acknowledge that despite the robustness of our uncertainty-aware learning (UAL) method in handling uncertainties, there are still challenges in situations with extreme occlusion or congestion. These scenarios can severely impact the quality of facial data and subsequently hinder accurate emotion classification. This limitation should be carefully considered, and future research may focus on developing enhanced techniques to improve emotion classification performance in such challenging conditions. 
Furthermore, it's crucial to acknowledge that our method may not be readily applicable to multi-modal group emotion tasks. This presents a key consideration for deploying group emotion analysis tasks in practical applications. Our commitment to uncertainty-aware learning in GER will continue, and we aspire to extend this methodology to other computer vision tasks. By addressing the critical aspect of uncertainty within the GER domain, our work contributes to the broader field of AI, facilitating the creation of more reliable and dependable AI agents in emotionally charged applications.


\footnotesize
\bibliographystyle{IEEEtran}
\normalem
\bibliography{IEEEtemplate}

\end{document}